\documentclass[letterpaper, 10 pt, conference]{ieeeconf}  

\IEEEoverridecommandlockouts                              

\usepackage[applemac]{inputenc}
\usepackage{graphicx}
\usepackage{amsfonts,amssymb,xfrac}
\usepackage{mathtools}
\usepackage{booktabs}
\usepackage{siunitx}
\usepackage{multirow}
\usepackage{hyperref}
\usepackage{cleveref}
\usepackage{arydshln}   

\usepackage{cite,soul,bm}
\usepackage[table,xcdraw]{xcolor}
\definecolor{bestgreen}{RGB}{212,235,212} 
\newcommand{\best}[1]{\cellcolor{bestgreen}\textbf{#1}}

\newtheorem{remark}{\textbf{Remark}}

\newif\ifshowcomments
        \showcommentsfalse 
    \ifshowcomments 
        \newcommand{\bae}[1]{\hl{[SB: #1]}\protect\color{black}} 
        \newcommand{\di}[1]{\hl{[DI: #1]}\protect\color{black}}
        \newcommand{\ft}[1]{\hl{[FT: #1]}\protect\color{black}} 
        \newcommand{\jd}[1]{\hl{[JD: #1]}\protect\color{black}}
        \newcommand{\avi}[1]{\hl{[AS: #1]}\color{black}}
    \else
        \newcommand{\bae}[1]{}
        \newcommand{\di}[1]{}
        \newcommand{\ft}[1]{}
        \newcommand{\jd}[1]{}
        \newcommand{\avi}[1]{}
    \fi

\newcommand{\sig}{\sigma}
\newcommand{\TTC}{\mathrm{TTC}}

\title{\LARGE \bf
SIGMO: Sigmoid Gated Interaction-Aware Motion Planning
}

\author{Avinash Singh$^1$, Manish Saroya$^1$, Fatemeh Cheraghi Pouria$^{1,2}$, Faizan Tariq$^1$, \\Jovin D'sa$^1$, David Isele$^1$, Yosuke Sakamoto$^1$, Sangjae Bae$^1$
\thanks{All work was done at HRI. $^1$Honda Research Institute, USA, Inc., San Jose, CA 95134, USA
        {\tt\small avinash\_singh,sbae@honda-ri.com}
        $^2$University of Illinois Urbana-Champaign, Urbana, IL 61801, USA.}%
 }

\begin{document}

\maketitle
\thispagestyle{empty}
\pagestyle{empty}

\begin{abstract}
Despite the widespread adoption of autonomous vehicle technologies, reliable navigation within cluttered environments remains arduous due to the necessity of interaction-aware strategies; example scenarios include: narrow passages with multiple agents, lane changing in dense traffic, and crossing unsignalized intersections. These scenarios often require highly proactive strategies, simultaneously demanding interpretability and quick adaptability, which challenge end-to-end and data-driven methods. An effective strategy requires: scalability to multiple agents, adaptability to changes in other agents' behavior, and harmony with surrounding traffic without freezing or becoming overly conservative. To meet these requirements, we present SIGMO, a sigmoid-gated interaction-aware motion planner, with two main contributions. First, we formulate polynomial sigmoid functions that analytically estimate yield scores of surrounding agents using physics-grounded interaction features that capture driving intuitions. 
Second, we integrate the polynomial sigmoid functions into a receding-horizon planner, which facilitates interaction-aware planning with arbitrary trajectory predictions. We validate SIGMO in interaction-intensive \jd{use highly interactive} simulations and hardware experiments. The results show that SIGMO balances proactive and defensive behavior while remaining computationally lightweight, serving as a strong baseline in a mobility stack. The simulation and hardware experiment videos are available here: 
\textit{\color{blue} https://sites.google.com/view/sigmoid-polygate}.

\end{abstract}

\section{INTRODUCTION}
Autonomous vehicles have progressed from research prototypes to commercial deployments, yet reliable navigation in cluttered, interaction-intensive environments remains challenging. Representative scenarios include traversing narrow passages shared with multiple agents, merging into dense traffic, and crossing unsignalized intersections. In such settings, classical robust planning is often insufficient because there may be no independently ``safe'' gap into which the ego vehicle can progress without interacting with other agents. Instead, the ego vehicle must \emph{negotiate}: proactively encouraging other agents to yield while remaining defensive when they do not. Overly conservative behavior can result in the well-known ``freezing robot'' problem \cite{trautman2010unfreezing}, whereas overly aggressive behavior can compromise safety. An effective planner must therefore (i) scale to multiple agents, (ii) adapt rapidly to changes in other agents' behavior, and (iii) remain compatible with surrounding traffic while providing interpretable decisions.

\begin{figure}
\centering
\includegraphics[width=0.8\columnwidth]{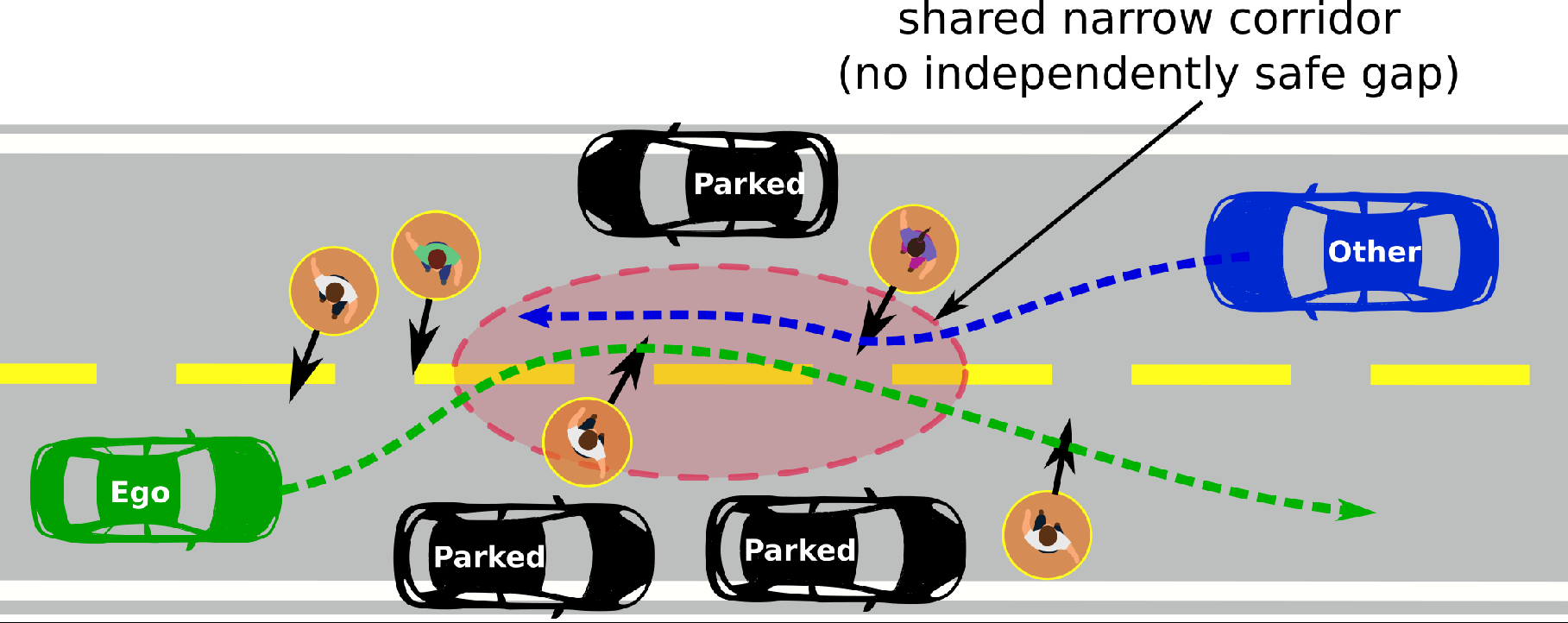}
\caption{\textbf{Motivating scenario}: navigating a narrow street where parked vehicles (black) force the ego vehicle and an oncoming agent (blue) into a shared corridor (red), while pedestrians move in different directions. No independently safe gap exists; the ego vehicle must determine whether the oncoming agent and pedestrians are likely to yield, proactively claim the corridor when appropriate, and defer otherwise.}
\label{fig:motivation}
\vspace{-3mm}
\end{figure}

A substantial body of work has investigated interaction-aware planning. Game-theoretic approaches \cite{isele2019interactive, sadigh2016planning, fisac2019hierarchical} explicitly model the mutual influence among agents, but their computational complexity can grow rapidly with the number of participants, limiting scalability. Learning-based methods, including end-to-end policies \cite{hu2023planning, jiang2023vad, zheng2024genad} and reinforcement learning (RL) \cite{saxena2019driving, bouton2019cooperation, toromanoff2020end}, are attractive for capturing complex interactive behaviors. However, limited interpretability, reliability, and robustness to distribution shift remain significant barriers to deployment in safety-critical settings \cite{garcia2015comprehensive, chen2024end}. Optimization-based planners augmented with neural interaction predictors, such as NNMPC \cite{bae2020cooperation, bae2022lane} and IANN-MPPI \cite{ryu2025iann}, provide another approach by explicitly evaluating how other agents may respond to candidate ego actions within a receding-horizon framework. Nevertheless, these methods require repeated inference of a prediction model, often a recurrent neural network, conditioned on individual candidate rollouts. Consequently, their computational cost grows with both the number of samples and the number of agents. Moreover, their learned interaction models may generalize poorly beyond the behaviors and agent types represented in the training data. Thus, a lightweight and interpretable mechanism for estimating an agent's willingness to yield, without repeated action-conditioned inference and without assuming a particular prediction model, is still needed.
\jd{i think we are short on interaction modeling references. we might get critisim that use of sigmoid is not new, co-operation estimation exisits and we have not cited closely related methods for pedestrain interaction. some ideas to address this:- cite papers like : social force model Helbing and Molnár (1995), Schwarting et. al (2019): social behgavior modeling and  Li and Tomizuka paper Interaction-Aware Behavior Planning that uses sigmoids. i would end thius paragraph with something like:- while some methods have used Logit/sigmoid functions to model yield probability, our contribution ius a multi-feature fusion hierarch of features along with intgration into a planning framework - not just a sigmoid function}

To address this gap, we present SIGMO, a sigmoid-gated interaction-aware motion planner, with two main contributions. \textbf{First}, we introduce \emph{polynomial sigmoid functions} (``Polygate'') that analytically map physics-grounded features 
(time-to-collision, recent acceleration, relative velocity, path priority, and body proximity) to the probability that a surrounding agent yields. The formulation captures intuitive driving behaviors through interpretable features, and its parameters, namely the inflection point and sharpness, have a direct physical interpretation and require minimal tuning. 
\textbf{Second}, we integrate Polygate into a sampling-based receding-horizon control framework, Model Predictive Path Integral (MPPI) control \cite{williams2017model}, where the estimated yield probability modulates an interaction-aware risk term. Polygate is evaluated from geometric features of each candidate rollout rather than from a prediction model conditioned on it, so it avoids the repeated inference required by prior approaches \cite{bae2020cooperation, ryu2025iann} and remains compatible with any choice of prediction model. This enables efficient interaction-aware planning while retaining an interpretable decision mechanism.
\jd{i modified this point becvause the previous othe implied causality of previous version didn't seem accurate}

We evaluate SIGMO in interaction-intensive simulated scenarios, including narrow-passage negotiation with pedestrian crossing and merging into dense traffic. We then conduct the generalizability analysis with MetaDrive \cite{li2022metadrive} initialized by Waymo Open Dataset \cite{ettinger2021large}. We demonstrate SIGMO with a 1/10 scaled F1Tenth racecar in a merging scenario. The results demonstrate that SIGMO balances proactive negotiation and defensive behavior in various scenes while maintaining computational efficiency. 


\section{POLYGATE: Sigmoid Functions Based Interaction Scoring}
As a preliminary, a sigmoid function $s(\cdot)$ projects a feature $x$ onto $[0,1]$ based on an inflection point $\mu$ and sharpness $\tau$: 
\begin{equation}
\sig(z) = \frac{1}{1+e^{-z}}, \qquad
s_x(x) = \sig\!\left(\pm\frac{x - \mu}{\tau}\right).
\label{eq:generic}
\end{equation}
The sign of the sigmoid function depends on the characteristic of each feature, and each feature is computed from the perspective of an agent (subscribed with $_a$) \textit{relative to} the ego vehicle (subscribed with $_e$); e.g., relative velocity feature reads $\Delta \mathbf v = \mathbf v_a - \mathbf v_e$.
\subsection{Intuition driven sigmoid functions}\label{sec:sigmoids}
The polynomial sigmoid function design begins with driving intuitions. An agent is more likely to yield against ego when: (i) a collision is anticipated, (ii) ego is aggressive, (iii) ego is faster in speed (in a way the gap is closing), and (iv) ego is closer to a target. 
These intuitions are fundamentally based on physics, which generalize. A sigmoid function is then respectively defined with a formal method. 
\subsubsection{\textbf{Collision Anticipation}} Time-to-collision (TTC) is an effective metric for estimating how imminent a collision is, when a current maneuver is continued -- either assuming constant maneuver or with predictions. 



Under constant relative velocity, contact occurs when
$\|\Delta\mathbf p + t\Delta\mathbf v\|^2 = r_c^2$,
where $\Delta\mathbf p=\mathbf p_a-\mathbf p_e$ and $r_c = r_{\mathrm{ego}} + r_{\mathrm{agent}}$
is the sum of the collision-circle radii.
Define $\alpha = \|\Delta\mathbf v\|^2$,
$\beta_q = 2\Delta\mathbf p\cdot\Delta\mathbf v$,
and $\zeta = \|\Delta\mathbf p\|^2-r_c^2$.
For $\alpha>\varepsilon$ and discriminant
$D=\beta_q^2-4\alpha\zeta\ge0$, the contact times are:
\begin{equation}
    t_{1,2}=\frac{-\beta_q\mp\sqrt{D}}{2\alpha}.
\end{equation}
The time-to-collision is $\TTC=\mathrm{clip}(t_1,0,T_{\max})$ \jd{what is Tmax - planning horizon?} \avi{it is TTC cutoff. not explicitly the planning horizon.}when these conditions hold and $t_2>0$, i.e., the contact interval has not already passed, and $\TTC=T_{\max}$ otherwise, denoting no collision in view. The TTC sigmoid reads:
\begin{equation}
s_{\TTC}(\TTC) = \sig\left(\frac{\TTC - \mu_{\TTC}}{\tau_{\TTC}}\right),
\end{equation}
where larger TTC values produce higher TTC scores, indicating other agent is not closing the gap
, corresponding to less imminent predicted collision.

Rather than assuming constant velocity, the planner has access to the full rolled-out ego trajectory $\{\mathbf p^{(i)}_{e,t}\}_{t=1}^T$ and the per-sample agent predictions $\{\mathbf p^{(i,m)}_{a,t}\}_{t=1}^T$, for sample $i$ and agent $m$. TTC is evaluated by walking these predicted trajectories: at each step~$t$, find the earliest future step $k \ge t$ at which the pairwise distance drops below $r_c$, and report the lead time $(k-t)\Delta t$:
\begin{equation}
\TTC_{i,t,m} =
\begin{cases}
(k^\star - t)\Delta t, &
  \text{if } k^\star 
  \text{ exists},\\
T_{\max}, & \text{otherwise,}
\end{cases}
\label{eq:ttc_walked}
\end{equation}
where $k^\star = \min\bigl\{k \ge t :
  \|\mathbf p^{(i,m)}_{a,k} - \mathbf p^{(i)}_{e,k}\| < r_c\bigr\}$. This naturally accommodates a wider range of predictions including non-linear behaviors  (e.g., turning or yielding agents).

\subsubsection{\textbf{Aggressiveness}}
The aggressiveness is featured by the sum of recent accelerations that represents: ``more acceleration is expected to be less negotiable.''; in other words, ``more braking is expected to be more negotiable''. Using the agent's logged longitudinal-acceleration history $\{a_k\}$:
\begin{equation}
\Sigma a = \sum_{k=t-K+1}^{t} a_k,\qquad K: \text{number of steps}.
\end{equation}
\jd{how do we pick K?}\avi{we work back from wanting to keep $\sim$1s history. so if we are sampling at 0.2s, then k=5. in rc cars @0.3s k=3}
Sign convention: $\Sigma a<0$ $\Rightarrow$ sustained braking over the
window; $\Sigma a>0$ $\Rightarrow$ sustained accelerating. The aggressiveness sigmoid then reads:
\begin{equation}
s_{\Sigma a}(\Sigma a) = \sig\left(-\frac{\Sigma a - \mu_{\Sigma a}}{\tau_{\Sigma a}}\right),
\end{equation}
with a high $\Sigma a$ (i.e., intense accelerations), it derives a low yield probability.



\subsubsection{\textbf{Relative Velocity}} The relative velocity from the current states implies a priority to a target point (when two agents are equally distant). We project the relative velocity onto the a vector from ego to agent, flip sign so that
``approaching'' is positive, and normalize by ego speed.
 Let the closing ratio be \jd{fix denominator} $c = -\dfrac{(\mathbf v_a-\mathbf v_e)\cdot \hat{\mathbf r}}{\|\mathbf v_e\ + \epsilon\|}$
with $\hat{\mathbf r}$ pointing from ego to agent. The relative velocity sigmoid reads: 
\begin{equation}
s_{\text{speed}}(c) = \sig\!\left(-\frac{c - \mu_{c}}{\tau_{c}}\right),
\end{equation}
with a high closing ratio $c > 0$ (i.e., fast approaching), it derives a low yield probability.


\subsubsection{\textbf{Path Priority}} The distance to a target point clearly indicates a priority (when two agents are heading to the same point); this generally applies to traffic scenarios such as highway merging, 4-way intersection, and others. 

Path priority, clipped to a symmetric range, reads:
\begin{equation}
\pi = \mathrm{clip}\!\Bigl(\|\mathbf g-\mathbf p_a\| -
\|\mathbf g-\mathbf p_e\|,\ -\pi_{\max},\ \pi_{\max}\Bigr).
\end{equation}
This formulation is encounter-agnostic
(works for head-on, same-direction, crossing, and detouring cases). \jd{how would we calcuate such goal for pedestrians? does it also include overtaking scenarios?}\avi{from my undestanding,  $g$ is the egos goal, rather than an estimated ped goal. @sangjae could you please clarify}\bae{Avi is correct.} The sigmoid function reads:
\begin{equation}
s_{\text{path}}(\pi) = \sig\!\left(\frac{\pi - \mu_{\pi}}{\tau_{\pi}}\right).
\end{equation}
$\pi>0$ means the ego reaches the conflict point first $\Rightarrow$ ego
has right-of-way; this acts as a soft gate so that a low-priority ego
cannot claim a yield even if TTC is large.

\subsubsection{\textbf{Proximity Gate}}
We additionally define a proximity gate, which imposes higher risk (in the sense of low yield probability), when agent is adjacent to ego -- this complements underestimated cases such as ego and agent driving in the same direction and speed toward a merging point; large TTC and neutral aggressiveness/relative velocity/path priority regardless of the adjancy of the two agents. In such case, the proximity gate suppresses a yield probability towards ego to encourage ego's protective behaviors.

We leverage an \emph{elliptical body-frame metric} \cite{khatib1986real,lavalle2006planning} rather
than Euclidean distance. The relative position $\Delta\mathbf p$ is projected into ego's heading frame and normalized by the vehicle half-extents $(L_h, W_h)$:
\begin{align}
d_\text{lon} &= |\Delta\mathbf p \cdot \hat{\mathbf f}_e|,
\qquad
d_\text{lat} = |\Delta\mathbf p \cdot \hat{\mathbf n}_e|, \\[4pt]
d_\text{body} &= \sqrt{\left(\frac{d_\text{lon}}{L_h}\right)^{\!2}
                      + \left(\frac{d_\text{lat}}{W_h}\right)^{\!2}},
\label{eq:dbody}
\end{align}
where $\hat{\mathbf f}_e$ denotes ego's longitudinal vector and $\hat{\mathbf n}_e$ denotes ego's normal vector.
The result is in \emph{body-radii units}: adjacent-lane traffic has
large $d_\text{lat}/W_h$ and therefore large $d_\text{body}$
($\approx 3$ for a typical $3\mathrm{m}$ lateral gap), while
same-lane following stays close ($\approx 1.2$ for a $3\mathrm{m}$
longitudinal gap). This makes the proximity gate naturally
direction-aware without any road-frame lane assumption.
The proximity gate reads:
\begin{equation}
s_{\text{prox}}(d_\text{body}) =
\sig\!\left(\frac{d_\text{body} - \mu_{d}}{\tau_{d}}\right).
\end{equation}
Close agents have $s_{\text{prox}}\!\to\!0$, preventing
same-direction car-following (high TTC, low closing ratio) from being misread as
``yielding''.


\subsection{Polygate Formulation}\label{sec:polygate}
Each respective sigmoid function (defined in Section~\ref{sec:sigmoids}) does not solely generalize and they could conflict each other. Thus we define ``Polygate" by fusing the five sigmoid functions in the following steps.
\subsubsection{TTC-gated interaction term} We first define an interaction term, as a convex combination of acceleration sum and relative velocity signals, normalized by the weights $w_{\Sigma a}, w_\text{speed}$:
\begin{equation}
I = \frac{w_{\Sigma a} s_{\Sigma a} + w_\text{speed} s_{\text{speed}}}{w_{\Sigma a} + w_\text{speed}}.
\end{equation} 

\subsubsection{TTC-weighted base yield}
The TTC score dominates when large; interaction term $I$ only matters when TTC is small
$(s_{\TTC}\!\to\!0)$, indicating ``interaction may not occur when two agents are unlikely to intersect'':
\begin{equation}
P_{\text{base}} = \frac{w_\text{TTC} s_{\TTC} + w_I(1 - s_{\TTC}) I}{w_\text{TTC}+w_I(1 - s_{\TTC})}.
\label{eq:pbase}
\end{equation}

\subsubsection{Path-priority blend}
Path priority provides an additive bias (``independent gate'' in softened form):
\begin{equation}
P' = \frac{w_\text{path}s_{\text{path}} + w_\text{base} P_{\text{base}}}{w_\text{path}+w_\text{base}}.
\end{equation}

\subsubsection{Proximity gate}
For close agents we blend toward a neutral prior $\bar P$:
\begin{equation}
P_{\text{yield}} = s_{\text{prox}} P' + (1 - s_{\text{prox}})\bar P.
\label{eq:pyield}
\end{equation}

\noindent When $s_{\text{prox}} \to 1$ (agent far in body-radii),
$P_\text{yield} \approx P'$. When $s_{\text{prox}} \to 0$ (close/overlapping),
$P_\text{yield} \to 0.3$ regardless of other features. 

The sigmoid parameters used in this paper are tabulated in \autoref{tab:sigmoid_params}. Note that the inflection point $\mu$ barely needs tuning -- since neutral points are straightforward -- except for $s_\text{prox}$ that adapts to a desired distance (based on ego size). We used the same params for our validations in sim and hardware demonstrations (except $\mu(d_\text{body})=0.5$).

\begin{table}[ht]
\centering
\caption{Sigmoid score parameters.}
\label{tab:sigmoid_params}
\begin{tabular}{l c c c c}
\toprule
Score & Feature & Sign & $\mu$ & $\tau$ \\
\midrule
$s_{\TTC}$         & TTC (s)                    & $+$ & 3.0  & 0.5 \\
$s_{\Sigma a}$   & $\Sigma a$ (m/s$^2$)       & $-$ & 0.0  & 1.0 \\
$s_{\text{speed}}$ & $c$ (ratio)                & $-$ & 0.0  & 1.0 \\
$s_{\text{path}}$  & $\pi$ (m)                  & $+$ & 0.0  & 3.0 \\
$s_{\text{prox}}$  & $d_{\text{body}}$ (body-radii) & $+$ & 4.0  & 0.5 \\
\midrule
\multicolumn{5}{l}{\textit{Polygate (combination) weights}} \\
\midrule
$w_{\Sigma a}$ & \multicolumn{3}{l}{Acceleration sum weight}  & 0.30 \\
$w_\text{speed}$ & \multicolumn{3}{l}{Closing-ratio weight}  & 0.40 \\
$w_\text{TTC}$ & \multicolumn{3}{l}{TTC weight}  & 0.80 \\
$w_I$ & \multicolumn{3}{l}{Interaction term weight}  & 0.20 \\
$w_\text{path}$ & \multicolumn{3}{l}{Path proximity weight}  & 0.10 \\
$w_\text{base}$ & \multicolumn{3}{l}{Base gate weight}  & 0.90 \\
$\bar{P}$ & \multicolumn{3}{l}{Proximity lower bound} & 0.30 \\
\bottomrule
\end{tabular}
\end{table}

\section{SIGMO: Polygated Interaction Aware Motion Planning}
\subsection{Primal Optimization problem}
We first define the primal problem in discrete time: \jd{does this mean if we have M agents then the nominal cost terms get added M times??? if not then we should move the agent sum inside} \avi{Moved agent sum inside}

\begin{align}
    \min_{u_{0:T-1}}\;&\sum_{k=0}^{T-1}\Bigl[J_\text{nom}(x_k,u_k,\mathbf g)
      + \sum_{m=1}^{M} J_\text{int}(x_k,\hat{x}_{k,m})\Bigr] \nonumber\\
      &\qquad + J_T(x_T,\mathbf g) \\
    \text{s.t.}\quad
    x_{k+1} &= f(x_k,u_k),\quad u_k\in\mathcal{U},\quad k=0,\ldots,T-1.\nonumber
\end{align}\label{eq:primal_opt}

where $J_\text{nom}$ collects the tracking and comfort terms of \autoref{table:default_cost}, $J_\text{int}$ collects the interaction-aware risk and collision penalties against agent $m$ (\autoref{sec:riskterm}, \autoref{sec:collision}), and $J_T$ is the terminal goal cost.

For $M$ surrounding agents and planning horizon $T$, where the system dynamics $f(x_k,u_k)$ represents bicycle kinematics \cite{kong2015kinematic}, $\mathcal{U}$ defines control bounds, and $\mathcal{X}$ defines feasible set for states. The estimated $\hat{x}_{k,m}$ denotes predictions for agent $m$ at step $k$ and $\mathbf g$ is the ego vehicle's local goal point \jd{array would mean a refence path and we dont have a path following cost defined}\avi{changed to local goal point only} of $\{x_\text{goal}, y_\text{goal}\}$. The cost function $J(\cdot)$ generically reads motion planning objectives, including a reference tracking, driving comfort, and risks, which are detailed in Section~\ref{sec:cost_composition}.

We solve this problem in \eqref{eq:primal_opt} with Model Predictive Path Integral (MPPI) \cite{williams2017model}, for its practicality in handling nonlinearity (in dynamics and constraints) and for its efficiency with parallel-computing structure. MPPI typically follows: (i) set a nominal control sequence, (ii) sample candidates around the nominal control sequence, (iii) evaluate the samples (parallelization applies this step), w.r.t., costs and constraints (iv) best candidate selection (with weighted sum of top candidates or single best). Details are referred to \cite{williams2017model,ryu2025iann}. 

\remark SIGMO incorporates interaction awareness in \textit{step iii}, using Polygate to modulate the spatial risk term, and therefore does not require agent predictions conditioned on each candidate ego trajectory unlike \cite{ryu2025iann, bae2020cooperation, bae2022lane}. 

\subsection{Cost Compositions}\label{sec:cost_composition}
\subsubsection{Complete Cost}
For $N$ samples of control sequences
$\{\mathbf u^{(i)}_{0:T-1}\}_{i=1}^{N}$, roll them through a bicycle
model \cite{kong2015kinematic}, and assign each sample $i$ a cost:
\begin{align}
J^{(i)} \;=\;& 
w_{y} J_{y\text{-goal}}^{(i)}
+ w_{\text{lat-}v} J_{\text{lat-}v}^{(i)}
+ w_{v} J_{\text{speed}}^{(i)}
+ w_{u} J_{\text{ctrl}}^{(i)}\nonumber\\
&+ w_{j} J_{\text{jerk}}^{(i)}
+ w_{\dot\delta} J_{\text{steer}}^{(i)}
+ w_{\text{lane}} J_{\text{lane}}^{(i)}
+ w_{\text{bnd}} J_{\text{bnd}}^{(i)}\nonumber\\
&+ w_{x} J_{\text{term-}x}^{(i)}
+ w_{\text{term}} J_{\text{term-}y}^{(i)}
+ \underbrace{\Bigl(
  w_{\text{int}} J_{\text{int}}^{(i)}
  + w_\text{col}J_{\text{col}}^{(i)}\Bigr)}_\text{risk term}.
\label{eq:totalcost}
\end{align}

The sigmoids are in the
\emph{interaction-aware risk term} $J_{\text{int}}^{(i)}$. 

\subsubsection{Nominal Cost -- reference tracking and comfort}
The nominal cost is generically formulated to encourage a smooth and target-oriented maneuver in position and speed, regardless of presence of other agents; tabulated in \autoref{table:default_cost}.

\begin{table}[h]
\centering
\caption{Nominal Cost Terms \jd{we dont mention any of the MPPI cost terms especially time horizon and number of samples which should be reported somewhere}}
\label{table:default_cost}
\footnotesize
\setlength{\tabcolsep}{3pt}
\renewcommand{\arraystretch}{1.35}   
\begin{tabular}{@{}lc@{}}
\hline
\textbf{Term (Symbol)} & \textbf{Definition} \\
\hline
Lateral goal tracking $J^{(i)}_{y\text{-goal}}$ &
  $\tfrac{1}{T}\sum_{t=1}^{T}(y_{i,t} - y_\text{goal})^2$ \\
Lateral velocity damping $J^{(i)}_{\text{lat-}v}$ &
  $\tfrac{1}{T}\sum_{t=1}^{T}(v_{i,t}\sin\theta_{i,t})^2$ \\
Speed deviation $J^{(i)}_\text{speed}$ &
  $\tfrac{1}{T}\sum_{t=1}^{T}(v_{i,t} - v_\text{des})^2$ \\
Control effort $J^{(i)}_\text{ctrl}$ &
  $\tfrac{1}{T}\sum_{t=1}^{T}(a_{i,t}^2 + \phi_{i,t}^2)$ \\
Jerk $J^{(i)}_\text{jerk}$ &
  $\tfrac{1}{T}\sum_{t=0}^{T-1}(\Delta a_{i,t})^2$ \\
Steering rate $J^{(i)}_\text{steer}$ &
  $\tfrac{1}{T}\sum_{t=0}^{T-1}(\Delta\phi_{i,t})^2$ \\
Lane keeping $J^{(i)}_\text{lane}$ &
  $\tfrac{1}{T}\sum_{t=1}^{T}\min\!\bigl(\min_l(y_{i,t}-c_l)^2,\,1\bigr)$ \\
Boundary violation $J^{(i)}_\text{bnd}$ &
  $\tfrac{1}{T}\sum_{t=1}^{T}\bigl[\ln\bigl(1+e^{-(d_{\text{sgn},i,t}-b)}\bigr)\bigr]^2$ \\
Terminal goal $J^{(i)}_{\text{term-}\xi},\ \xi\!\in\!\{x,y\}$ &
  $|\xi_\text{goal}-\xi_{i,T}|\cdot\mathbf{1}\bigl[|\xi_{i,T}-\xi_\text{goal}|\ge\epsilon\bigr]$ \\
\hline
\end{tabular}
\end{table}

While most terms are self-explanatory and self-sufficient, lane keeping and boundary violation terms demand additional inputs. The lane keeping term requires lane centerlines $\{c_l\}_{l=1}^L$ for all lane $L$ if applicable. The boundary violation term requires boundary geometries $b = \{y_\text{min}, y_\text{max}\}$ along with signed interior distance
$d_\text{sgn} = \min(y - y_\text{min},\; y_\text{max} - y)$. The centerlines and boundary terms are substitutional.

\subsubsection{Interaction-aware Risk}
\label{sec:riskterm}

\begin{figure}
    \centering
    \includegraphics[width=0.8\columnwidth]{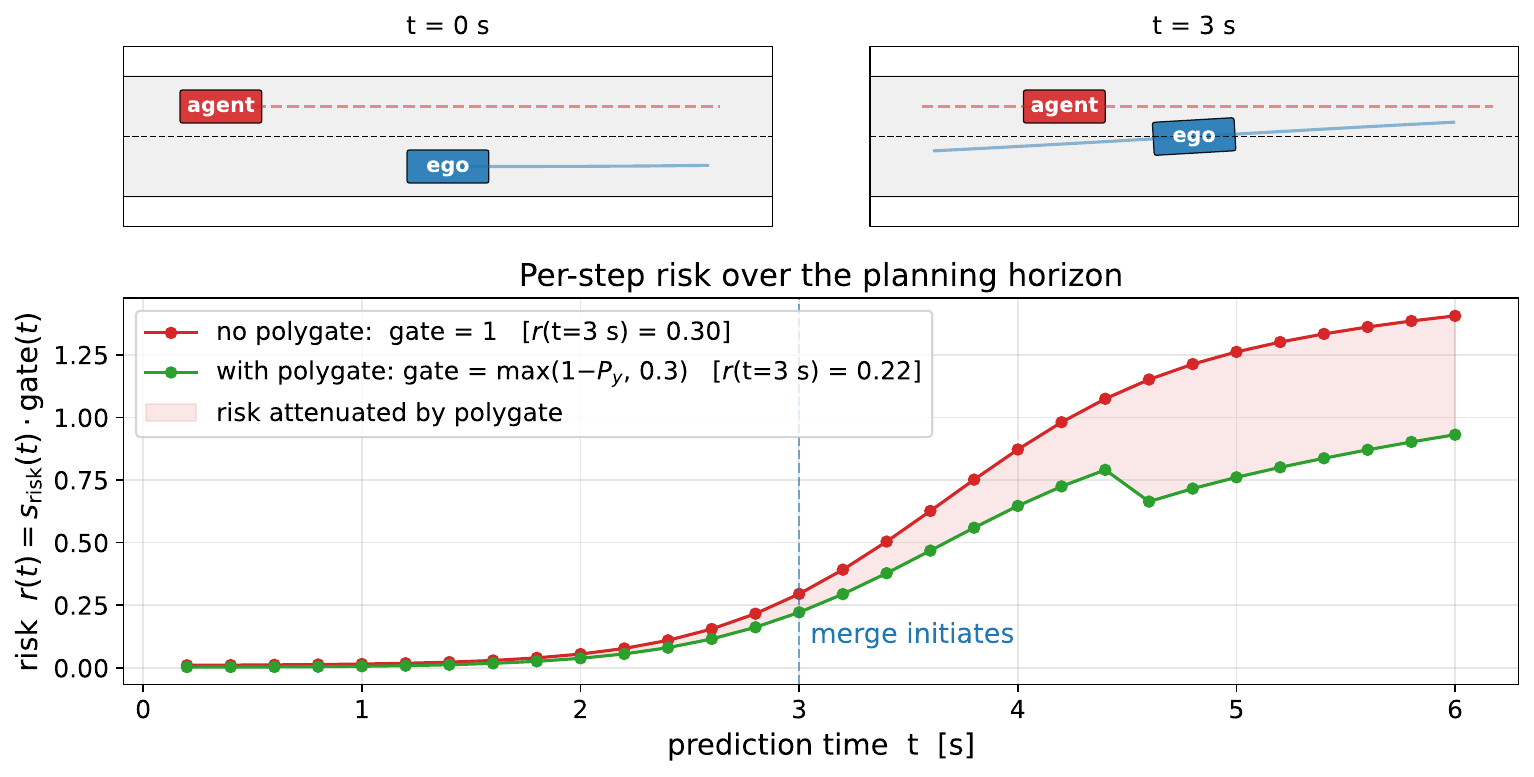}
    \vspace{-3mm}
    \caption{Risks with and without polygate in a merge scenario. Interaction-aware risk in \eqref{eq:riskcost} essentially diminishes along with the yield probability of the agent (red) to the ego vehicle (blue).}
    \label{fig:risk_demo}
    \vspace{-3mm}
\end{figure}

For sample $i$, prediction step $t=1,\ldots,T$, and agent
$m=1,\ldots,M$ within interaction radius $R_{\text{int}}$, the risk estimation $\mathrm{risk}^{(i,m)}_t$ is:
\begin{equation}
\mathrm{risk}^{(i,m)}_t \;=\;
\underbrace{\gamma^{t}}_{\text{decay}}
\cdot\underbrace{\mathcal B\!\left(\mathbf p^{(i)}_t,\hat{\mathbf p}^{m}_t;\theta\right)}_{\text{spatial risk}}
\cdot
\underbrace{g\left(P_{\text{yield}}^{(i,m,t)}\right)}_{\text{cooperation gate}},
\label{eq:intriskterm}
\end{equation}
where the spatial risk $\mathcal B(\cdot)$ represents the Bhattacharyya-kernel overlap \cite{bhattacharyya1946measure, tariq2023rcms}
between two oriented 2D Gaussians with extents
$(\ell_{\text{cov}},w_{\text{cov}})$, producing high value when the two Gaussians heavily overlap. The polygate yield probability from~\eqref{eq:pyield} acts as a \emph{risk multiplier}:
\begin{equation}
g(P_{\text{yield}}) \;=\; \max\!\bigl(1 - P_{\text{yield}},\; 0.3\bigr).
\label{eq:coopgate}
\end{equation}
With the cooperation gate, the interaction-aware risk says ``Heavily consider the risk associated with agent $m$, as the agent $m$ gets non-cooperative and vice versa''. 
This gated risk promotes interaction awareness, by imposing priority and importance against each agent. 
Note that the risk multiplier is lower-bounded --
risk is attenuated but never eliminated, since a yielding
agent is still a physical obstacle. Summed over time and agents, the interaction-aware risk term reads:
\begin{equation}
J_{\text{risk}}^{(i)} \;=\;
\sum_{t=1}^{T}\sum_{m=1}^{M}
\mathrm{risk}^{(i,m)}_t.
\label{eq:riskcost}
\end{equation}

\autoref{fig:risk_demo} illustrates how the interaction-aware risk diminishes with polygate in a merging scenario. The polygate imposes priority and accordingly attenuates risks, allowing the ego vehicle to merge.


\subsubsection{Collision Risk}
\label{sec:collision}

Because MPPI imposes no hard constraints, we add a dominating collision penalty, restricted to the near horizon $T_c=\min(T,1/\Delta t)$, i.e., one second, to avoid over-conservative behavior in interactive scenes.

Extending \cite{ryu2025iann} which applied three circles, each agent is represented by one or more circles based on its size $(\ell, w)=(2\ell_h, 2w_h)$. The number of circles is computed as
$\max\bigl(1,\lceil\ell / w\rceil\bigr)$; e.g., $\ell=5\mathrm{m},w=2\mathrm{m} \rightarrow 3$ circles. For ego circle $p$ and agent $m$ circle $q$, the margin is $\delta_{p,q} = \|\mathbf c_p^\text{ego} - \mathbf c_q^{(m)}\| - r_\text{comb}$, where $r_\mathrm{comb} = w_h^\text{ego}+w_h^\text{agent}$. The hard collision penalty explains overlaps between any pair of circles, formally written as: 
\begin{equation}
J^{(i)}_\text{hard} = \alpha_\text{hard}\sum_{t=1}^{T_c}\sum_{m=1}^{M}
\Bigl[\max\!\bigl(-\min_{p,q}\delta_{p,q},\;0\bigr)\Bigr]^2.
\end{equation}

To facilitate continuous risk estimations against collision, The ego-agent gap is projected into the ego body frame and normalized by
anisotropic safe extents
$S_\text{lon} = r_\text{comb}\alpha_\text{lon}$ and
$S_\text{lat} = r_\text{comb}\alpha_\text{lat}$:
\begin{equation}
d_\text{norm} = \sqrt{\left(\frac{d_\text{lon}}{S_\text{lon}}\right)^{\!2}
               + \left(\frac{d_\text{lat}}{S_\text{lat}}\right)^{\!2}\;}.
\end{equation}
The elongated longitudinal extent (i.e.,, $\alpha_\text{lon} > \alpha_\text{lat}$) provides a deeper warning for
following gaps while being fairly neglectful of adjacent-lane traffic. The soft collision penalty reads:
\begin{equation}
J^{(i)}_\text{soft} = \alpha_\text{soft}\sum_{t=1}^{T_c}\sum_{m=1}^{M}
\Bigl[\max\!\bigl(1 - d_\text{norm},\;0\bigr)\Bigr]^2,
\end{equation}
with the scaler $\alpha_\text{soft}$ is set to reasonably attenuate collision weight $w_\text{col}$; e.g., $\alpha_\text{soft}=0.2$ in our experiments. 

The final collision penalty is the sum of the hard and soft collision term:
\begin{equation}
J^{(i)}_\text{col} \;=\; J^{(i)}_\text{hard} + J^{(i)}_\text{soft}.
\end{equation}
\remark 
The collision penalties remain independent of Polygate: the hard term penalizes circle overlap, while the soft term penalizes approach within the safe extents $d_\text{norm}$.

\section{Validations}
\subsection{Simulations}
\subsubsection{Scenarios}
We develop simulated environments that have interaction-intensive scenarios. We consider two scenarios: (i) \textbf{narrow passage} with pedestrians and other vehicles and, (ii) \textbf{merge} into dense traffic. While both are interaction-intensive, each poses distinct challenges. In the narrow passage scenario, the ego vehicle must pass pedestrians when possible, judge when to slow down and when to speed up, and maintain safety primarily through braking. In contrast, in the merge scenario, it must nudge an adjacent vehicle into yielding (i.e., pro-activeness) as it approaches the target lane, balance maintaining a target speed with completing the merge sooner, and maintain safety primarily by steering away when the adjacent vehicle refuses to yield. In the merge scenario, we set the desired speed $v_\text{des}$ to the ego vehicle's current speed, which reduces the incentive to gain position through acceleration. 

\subsubsection{Behavioral modeling}\label{sec:behavior_modeling}
We use responsive behavioral models for other agents to evaluate how the planner adapts to different agent behaviors. Both models have a configurable cooperativeness parameter that is hidden from the planner.

\textbf{Pedestrians.} In the narrow passage scenario, each pedestrian $i$ is modeled as a point mass with state $\mathbf{s}_i=[x_i,y_i,v_{x,i},v_{y,i}]^{\top}$ and a type-conditioned, goal-directed policy parameterized by a latent cooperativeness variable $c_i\in[0,1]$. We consider four pedestrian types, $\tau_i\in\{\textsc{same\_dir},\textsc{oncoming},\textsc{crossing},\textsc{loitering}\}$. For non-loitering pedestrian $i$, the desired velocity is $\mathbf{v}^{\text{des}}_i=(1-c_i\rho_i)v_i^{\text{pref}}\hat{\mathbf{g}}_i$, where $\hat{\mathbf{g}}_i$ points toward the sampled goal and $\rho_i=\max(0,1-\|\mathbf{p}_i-\mathbf{p}_{\text{ego}}\|/d_{\text{react}})$ measures ego proximity with $d_{\text{react}}=8\,\mathrm{m}$, with trajectory $\mathbf{p}_{(\cdot)}$. 
Thus, cooperative pedestrians progressively reduce their speed as the ego approaches and, when $c_i>0.3$ and $\rho_i>0.3$, bias their heading toward the nearer inner sidewalk to model lateral yielding. Loitering pedestrians instead undergo bounded stochastic drift with Gaussian acceleration noise $\sigma_\ell=0.3\,\mathrm{m/s}^2$ and damping $\lambda_\ell=0.5\,\mathrm{s}^{-1}$, with cooperative loiterers additionally attracted toward the nearer sidewalk when the ego is nearby. Independently of $c_i$, all pedestrians apply a reactive lateral repulsion when the free gap $d_{ij}$ to a nearby vehicle $j$ falls below $d_{\text{avoid}}=8\,\mathrm{m}$, using $a_{y,i}^{\text{avoid}}=k_{\text{avoid}}\max(0,1-d_{ij}/d_{\text{avoid}})\operatorname{sign}(y_i-y_j)$ with $k_{\text{avoid}}=2\,\mathrm{m/s}^2$; no pedestrian is therefore fully collision-ignorant. The resulting acceleration is $\mathbf{a}_i=(\mathbf{v}^{\text{des}}_i-\mathbf{v}_i)/\tau_v+\bm{\epsilon}_i+a_{y,i}^{\text{avoid}}\mathbf{e}_y$, with $\tau_v=0.5\,\mathrm{s}$ and $\bm{\epsilon}_i\sim\mathcal{N}(\mathbf{0},0.08^2\mathbf{I})$, and the state is integrated using Euler steps of $\Delta t=0.2\,\mathrm{s}$ subject to corridor-wall constraints. This model separates pedestrian intent, captured by $(\mathbf{g}_i,\tau_i)$, from disposition, captured by the latent variable $c_i$, enabling the ego planner, whether SIGMO or a baseline, to infer cooperativeness from observed motion, which provides a controlled mechanism for evaluating interaction-aware planning.

\textbf{Vehicles.} In the merge scenario, surrounding vehicles follow the interactive behavior models of \cite{bae2020cooperation, ryu2025iann}. Each vehicle tracks its lane with the Intelligent Driver Model (IDM) \cite{treiber2000congested}, which on its own reacts only to the vehicle ahead in the same lane. The model is extended to add a yielding decision toward a merging vehicle from the adjacent lane. When the ego vehicle approaches the lane boundary, each surrounding vehicle decides whether to treat it as in-lane, and hence yield, by a Bernoulli draw with parameter $p_m$. A high value of $p_m$ yields readily, whereas a low value of it represents more aggressive behavior where the surrounding vehicle ignores the ego. Please refer to \cite{bae2020cooperation} for details of the underlying interaction model.


\begin{figure}
    \centering
    \includegraphics[width=0.9\columnwidth]{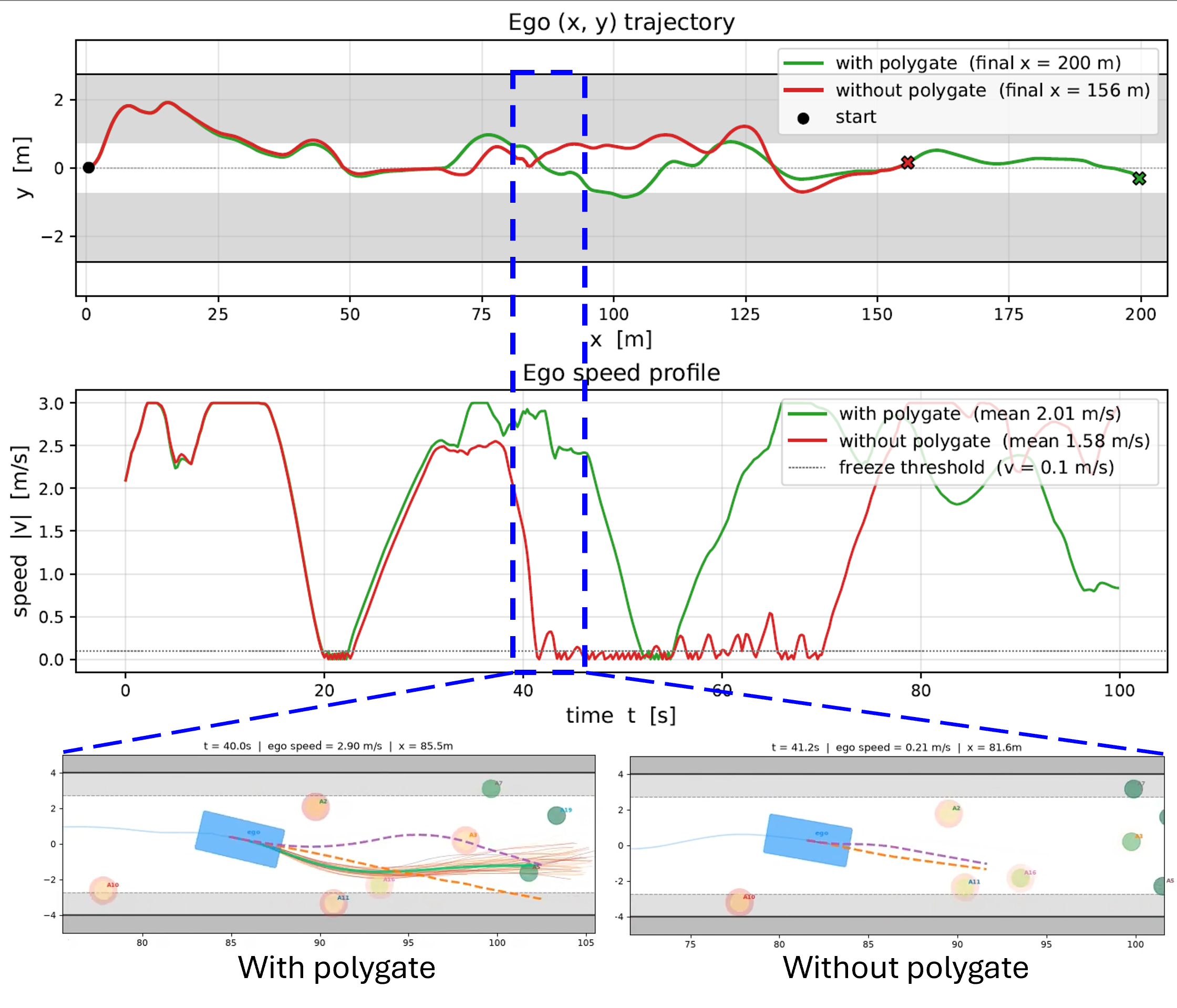}
    \caption{SIGMO trajectory in \textbf{narrow passage} scenario. SIGMO with the polygate estimates the responsiveness of other agents, successfully navigating through a cluttered corridor. In contrast, the absence of polygate purely relies on the constant prediction, stranding the ego vehicle in the cluttered corridor. The color of an agent indicates the estimated risk to the ego vehicle (i.e., red$\uparrow$ $\rightarrow$ risk$\uparrow$).}
    \label{fig:narrow_passage_qual}
    \vspace{-3mm}
\end{figure}

\begin{figure}
    \centering
    \includegraphics[width=0.9\linewidth]{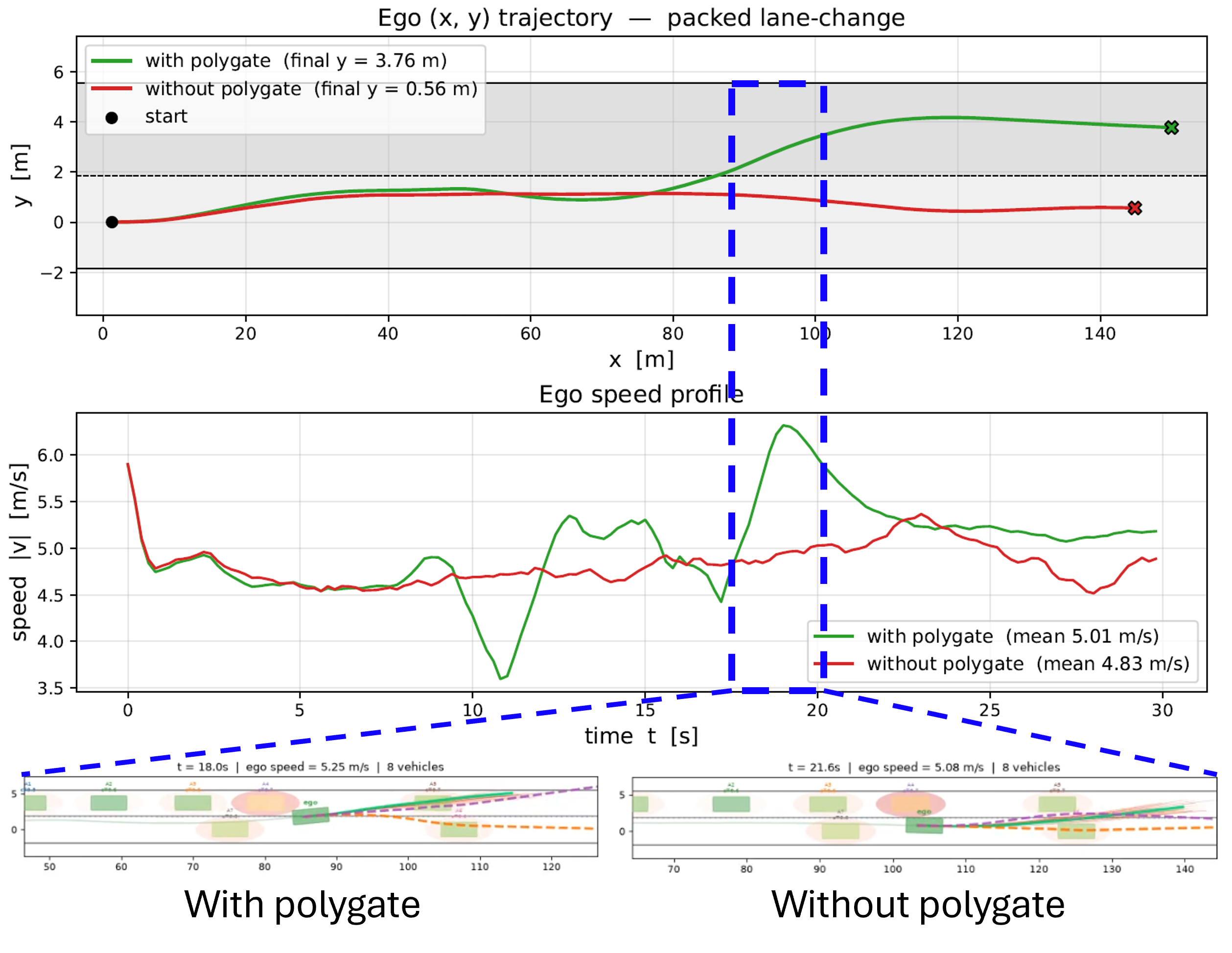}
    \caption{SIGMO trajectory in \textbf{merge} scenario. SIGMO with the polygate estimates yield probability of the agent in the target lane, successfully making space and merging. In contrast, the pure constant prediction does not solely predict a yield behavior, having the ego vehicle remain in the source lane. The color of an agent indicates the estimated risk to the ego vehicle (i.e., red$\uparrow$ $\rightarrow$ risk$\uparrow$).}
    \label{fig:merge_qual}
\end{figure}

\begin{figure}
    \centering
    \includegraphics[width=0.8\linewidth]{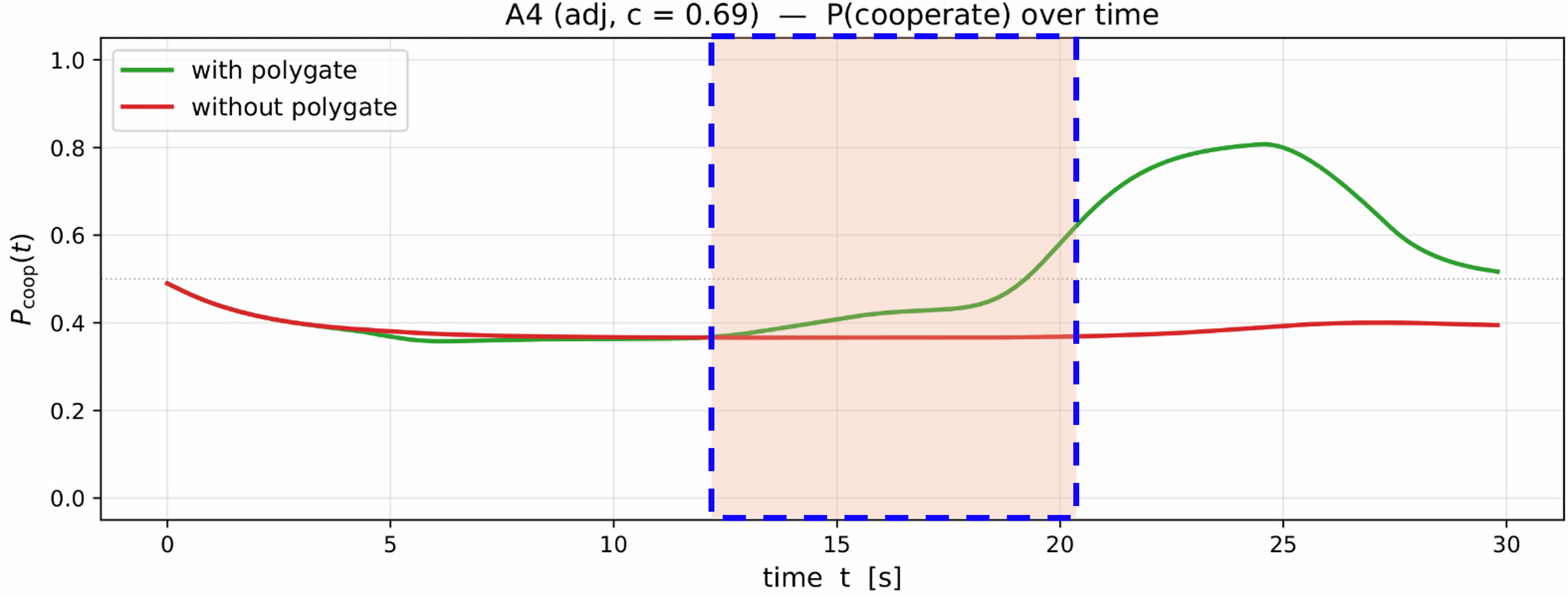}
    \caption{Yield probability estimation in \textbf{merge} scenario. Agent A4 is the key agent that the ego vehicle interacts with to make space to merge. SIGMO with the polygate nudges A4 to slow down, and the yield probability increases as A4 decelerates, leading the ego vehicle to decide to merge.}
    \label{fig:yield_prob_merge}
    \vspace{-3mm}
\end{figure}

\subsubsection{Baselines for Comparative Analysis}\label{sec:baselines}
We first compare SIGMO against the base planner and two oracles. (i) \textbf{MPPI} has no additional interaction-aware risk term apart from the collision penalties of Section~\ref{sec:collision}, which act over the next $T_c = 1\,\mathrm{s}$. (ii) \textbf{Oracle-yield} replaces the estimated $P_\text{yield}$ with each agent's true cooperativeness parameter ($c_i$ or $p_m$) while keeping the constant-velocity prediction. (iii) \textbf{Oracle-traj} replaces the constant-velocity prediction with each agent's actual future trajectory, obtained by rolling the simulator forward under every candidate ego trajectory. Oracle-yield bounds what a perfect yield estimate can achieve with the same prediction, and Oracle-traj bounds what a perfect, reactive prediction can achieve.


\subsubsection{Metrics and protocol}\label{sec:metrics}
Each scenario is run for 100 seeds; a seed fixes the initial positions, speeds, and cooperativeness parameters of all agents. A narrow-passage run succeeds if the ego reaches the end of the corridor within 90\,seconds without collision; a merge run succeeds if it reaches the target lane within 30\,seconds without collision. Runs that neither succeed nor collide are timeouts. We also report the minimum ego--agent distance, completion time over successful runs, and, for the narrow passage, stall time: the time spent below 1\,m/s, averaged over runs in which the ego stalled.

\subsection{Qualitative Analysis}
We first examine a narrow passage scenario shared by several pedestrians and cars. The simulated pedestrians nominally are prioritized, however, they routinely make room for a vehicle sharing the corridor by stepping aside and/or stopping/creeping for a car to pass. They can potentially switch immediate decisions based on other agents' behaviors (e.g., if an agent is aggressively approaching, a pedestrian may decide to yield). These behaviors require the ego vehicle to adjust its speed based on cooperation. \autoref{fig:narrow_passage_qual} illustrates the performance of SIGMO compared to a naive MPPI with constant velocity (i.e., without the polygate in Eqn.~\eqref{sec:polygate}). Around $t=40$, the ego vehicle approaches pedestrians crossing the corridor. SIGMO accelerates as it approaches these pedestrians, such that crossing pedestrians would yield.  At around $t=50$ the ego vehicle decides to slow down for an oncoming pedestrian whose path overlaps its own. The same planner produces both behaviors showcasing the timely and interchangeable strategies to navigate through the crowd. In contrast, naive MPPI, which reacts to a pedestrian only when a collision is within one second, brakes late and repeatedly as pedestrians cross, resulting in being stranded for an extended time period (more than 10$\mathrm{s}$). 

The merge scenario is simpler compared to the narrow passage scenario since vehicle behaviors can be characterized in a straightforward manner: each vehicle either opens a gap in front or closes it. However, the merge scenario very clearly presents a ``proactive maneuver'':with no gap available, the ego must nudge a target-lane vehicle into making one -- and we focus on the qualitative analysis on the nudge. \autoref{fig:merge_qual} illustrates this interaction. Both the ego vehicle and other vehicles on the target lane drive at similar speeds, providing little relative motion through which an existing gap could be leveraged. SIGMO (i.e., with polygate) consistently presents the proactive merging mechanism: (i) it slowly approaches the vehicle in the target lane; (ii) if the vehicle decides to slow down, the ego merges in front of the yielding vehicle; (iii) if the vehicle holds its speed, the ego repeats the nudge at the next time step; (iv) if the vehicle decides to speed up, the ego tries to merge behind the vehicle. In contrast, the naive MPPI with constant prediction (i.e., without polygate) and no interaction term, ends up staying in the source lane. \autoref{fig:yield_prob_merge} shows how the yield estimate varies over time during the interaction. As the target-lane vehicle decelerates, its estimated yield probability increases. This can reduce the corresponding interaction-risk multiplier, encouraging itself to behave more proactively to merge, which is not observed in the naive MPPI (without polygate). 


\begin{figure*}
    \centering
    \includegraphics[
        width=0.9\linewidth,
        trim={0cm 0.5cm 0cm 1.5cm},
        clip
    ]{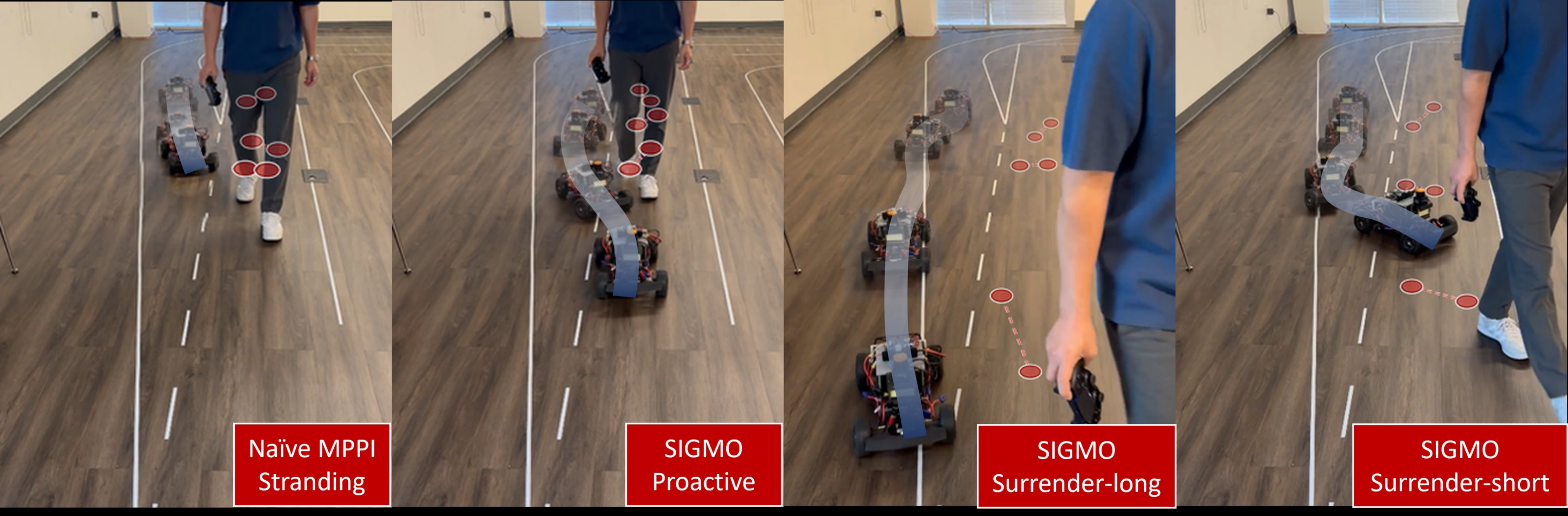}
    \vspace{-3mm}
    \caption{Hardware experiments with F1Tenth \cite{o2020f1tenth}. The ego vehicle interacts with the pedestrian to merge from the ramp. The first image is with Naive MPPI and the rest are with SIGMO (showcasing various behaviors). All the experiments are with the same hardware settings (e.g., localization and perception) as well as the pedestrian prediction. Each pair of red circles indicates the footprints of the pedestrian.}
    \label{fig:hardware-experiments}
    \vspace{-3mm}
\end{figure*}

\subsection{Comparative Analysis}
We now quantify these behaviors over the 100-seed protocol of Section~\ref{sec:metrics}, comparing against the baselines of Section~\ref{sec:baselines}; \autoref{tab:mc_polygate} reports the results. 

In the narrow passage scenario, \textbf{SIGMO} adds 4 percent to the success rate compared to \textbf{MPPI}. A few interesting points of discussion: (i) the improvement in success rate is not significant, however, the stall time is significantly (24.3\%) reduced; (ii) at the same time, \textbf{SIGMO} does not harm safety \jd{are these at-fault collision? if yes let's clarify because 25\% collision looks bad and makes interaction awarenes questionable}(having a similar collision rate and minimum distance), only benefiting performance -- the current collisions are primarily not-at-fault; (iii) Interestingly, \textbf{SIGMO} has a higher success rate and fewer collisions than \textbf{Oracle-yield}, which has a true cooperation parameter for traffic agents; it can be seen that the residual in the yield estimate leads to more proactive behaviors, resulting in more successful and safer maneuvers. (iv) The performance can further improve by having ``perfect prediction on response'' (i.e., \textbf{Oracle-traj}), yet the success rate stays at 93\%, indicating the difficulty of the test case. 

In the merge scenario, \textbf{SIGMO} significantly improves the success rate ($\uparrow$12\%), as this scenario often requires a heavy nudge to merge, which \textbf{MPPI} does not perform. \textbf{SIGMO} does not harm collisions (consistent with the narrow passage scenario), again showcasing upsides without downsides; it converts timeout to success. That said, the minimum distance decreases (0.2$\mathrm{m}$), and this is rooted in the nudging behavior that closes lateral distance to the adjacent agent. The completion time (i.e., time-to-merge) also benefits from \textbf{SIGMO}; while the number does not seem significant (1.5$\mathrm{s}$), the completion time is lower-bounded to 19$\mathrm{s}$ with a perfect prediction, indicating 36.5\% performance closing. The success rate in this scenario is capped at 69\%, which indicates the fundamental infeasibility of merging.



  \begin{table}[t]
  \centering
  \caption{Monte-Carlo evaluation ($N{=}100$ each).}
  \label{tab:mc_polygate}
  \setlength{\tabcolsep}{2.5pt}
  \renewcommand{\arraystretch}{1.05}
  \scriptsize
  \begin{tabular}{@{}l r r r r r r@{}}
  \toprule
  Model & Succ. & Coll. & T.O. & Min-d & $\bar{t}_\text{comp}$ & Stall [s] \\
        & [\%]  & [\%]  & [\%] & [$\mathrm{m}$]   & [s]                   &           \\
  \midrule
  \multicolumn{7}{@{}l}{\emph{Narrow passage}} \\
  \textbf{MPPI}        & 69        & 26        & 5         & \best{0.37} & 71.8        & 18.1 (47) \\
  \rowcolor{red!8}
  \textbf{SIGMO (ours)}        & \best{73} & \best{25} & \best{2}  & 0.36        & \best{70.6} & \best{13.7 (43)} \\
  \hdashline[1pt/2pt]\noalign{\vskip 1pt}
  \textbf{Oracle-yield} (yield est. bound) & 70        & 30        & 0         & 0.36        & 69.0        & 11.3 (36) \\
  \textbf{Oracle-traj} (performance bound) & 93        & 7         & 0         & 0.57        & 66.5        & 8.2 (17) \\
  \midrule
  \multicolumn{7}{@{}l}{\emph{Merge (packed)}} \\
  \textbf{MPPI}        & 26        & \best{2}  & 72        & \best{0.82} & 23.1        & --- \\
  \rowcolor{red!8}
  \textbf{SIGMO (ours)}        & \best{38} & \best{2}  & \best{60} & 0.62        & \best{21.6} & --- \\
  \hdashline[1pt/2pt]\noalign{\vskip 1pt}
  \textbf{Oracle-yield} (yield est. bound) & 49        & 1         & 50        & 0.51        & 20.3        & --- \\
  \textbf{Oracle-traj} (performance bound)  & 69        & 6         & 25        & 0.26        & 19.0        & --- \\
  \bottomrule
  \end{tabular}
  \vspace{-3mm}
\end{table}

\subsection{Generalizability Analysis with MetaDrive}
To test the generalizability of SIGMO (in various scenes), we further extend a quantitative analysis in MetaDrive \cite{li2022metadrive} with one of the benchmark interaction-aware planners, IANN-MPPI \cite{ryu2025iann}; it has an MPPI backbone like ours, but the interaction is evaluated by a pre-trained Neural Network (Social GAN \cite{gupta2018social}). \autoref{tab:quantitative_results} reports the performance with 150 scenarios, each of which is initialized with the Waymo Open Dataset \cite{ettinger2021large}; The scenario set comprises diverse driving environments (including left turn, right turn, lane-following, lane-change, T-intersections, 4-way intersection, and highway merge) that enable evaluation across a wide range of traffic conditions and road layouts. The results present a superior performance of SIGMO, indicating a more generalizable interaction-estimation compared to IANN-MPPI that was trained on a specific case (merge scenario). The collisions in both models primarily occur in high-speed scenarios and merging; meanwhile, SIGMO brings 3.4\% of collisions (with IANN-MPPI) to success. 

\begin{table}[t]
  \centering
  \caption{Performance comparison in MetaDrive on 150 evaluation scenarios. OOR denotes Out-of-Road events.\jd{can we add avg runtime if already available since we claim lightweight?}\bae{good point, if we have the numbers, please add, otherwise, we discussed the efficiency in the RC result section.}}
  \label{tab:quantitative_results}
  \small
  \setlength{\tabcolsep}{3.5pt}          
  \begin{tabular}{@{}l c c c c@{}}       
  \toprule
  Method & Succ.\,[\%] & Coll.\,[\%] & OOR\,[\%] & T.O.\,[\%] \\
  \midrule
  SIGMO (ours)        & \textbf{92.7} & \textbf{7.3} & 0.0 & 0.0 \\
  IANN-MPPI~\cite{ryu2025iann} & 89.3       & 10.7          & 0.0 & 0.0 \\
  \bottomrule
  \end{tabular}
  \vspace{-3mm}
\end{table}


\subsection{F1Tenth Racing Car Experiments \& Efficiency}
We demonstrate the performance of SIGMO on a 1/10th scale F1Tenth racecar \cite{o2020f1tenth}, in an indoor track with a ramp and merging lane. The test car is equipped with 
an Asus NUC 15 Pro with Intel ARC GPU/AI-ready NPU.
SIGMO runs in parallel with the perception (Hokuyo 10LX Lidar) and localization modules, requiring efficient real-time operation. SIGMO maintains an average processing frequency of 15Hz, with MPPI parameters of \{2000 samples, 2 optimization iterations\}, and a planning horizon of 10 steps. 

\autoref{fig:hardware-experiments} presents three experiments of SIGMO in comparison to an experiment of Naive MPPI as a baseline. The first two illustrate the same scenario for both MPPI and SIGMO, where MPPI fails to merge, as opposed to SIGMO. The finding consistently holds: the polygate effectively attenuates risks along the predicted path of the pedestrian, as a result of yield likelihood estimation. In contrast, MPPI naively relies on the prediction of the pedestrian, limiting itself to a reactive maneuver. The right two examples show ``Surrender'' case of SIGMO: In ``Surrender-long'' case, SIGMO consistently tries to overtake but ends up surrendering and merging behind the pedestrian. In ``Surrender-short'' case, SIGMO tries to overtake during a short time period, and gives up quickly, merging behind the pedestrian. The results illustrate diversity in SIGMO's behavior, with timely decisions on surrender.




\subsection{Limitations and future works}
The limitation of this paper primarily lies in traffic interaction models being relatively simple (Section~\ref{sec:behavior_modeling}); which may not represent realistic human behaviors. Thus, future work includes more rigorous simulation settings with more realistic and complex interactive behaviors. 
Another limitation is that SIGMO is built upon MPPI, and thus it shares the limitation of not having hard constraints for collision avoidance. That being said, the polygate in SIGMO is not limited to the MPPI scheme, and it can be easily integrated into other schemes (as an additive penalty) that have hard constraints, e.g., model predictive control. \jd{we could add hand selected weights? do we have ablations for each sigmoid interaction term to say all were needed and leaving one out reduced performance? if yes let's add that to the table}\bae{That's a good point. We do not have ablations, unfortunately. But in fact, we used the same weights of sigmoids for both sim and experiments.}

\section{CONCLUSIONS}
This paper presents principled formulations and validations of an interaction-aware motion planning scheme; named as SIGMO (Sigmoids Gated Interaction-Aware Motion Planning). In particular, this paper first formalizes yield probability estimate by combining five sigmoid functions that represent physical correlations between two agents. As combined, the sigmoid functions are integrated into a MPPI scheme as a risk multiplier; adding interaction-awareness to a canonical planner even with a naive non-interactive prediction (e.g., constant prediction). The validations in both simulations and hardware experiments support the potential of the proposed scheme for achieving proactive and interaction-aware planning for autonomous driving. 

\section*{Acknowledgement}
We acknowledge the use of Claude (https://claude.ai) to assist with setting up the simulation environment, structuring tables for reporting used parameters and results, and generating figures. The authors reviewed, edited, and take full responsibility for the content and integrity of this manuscript.




\bibliographystyle{IEEEtran}
\bibliography{refs}

\end{document}